\documentclass[letterpaper]{article} 
\usepackage[preprint]{aaai2027}
\usepackage[hyphens]{url}  
\usepackage{graphicx} 
\usepackage{natbib}  
\usepackage{caption} 
\usepackage{xcolor}
\usepackage{booktabs}
\usepackage{tabularx}
\usepackage{array}
\usepackage{makecell}

\definecolor{codegray}{gray}{0.93}
\newsavebox{\promptbox}
\newcolumntype{Y}{>{\raggedright\arraybackslash}X}

\usepackage{tcolorbox}
\tcbuselibrary{skins}
\usepackage{algorithm}
\usepackage{algorithmic}

\author{
    Haodong Chen\textsuperscript{\rm 1}\equalcontrib,
    Yadong Wang\textsuperscript{\rm 1}\equalcontrib,
    Shengtao Wen\textsuperscript{\rm 1},
    Dong Liang\textsuperscript{\rm 1},
    Xiang Chen\textsuperscript{\rm 1}\thanks{Corresponding author.}
}
\affiliations{
    \textsuperscript{\rm 1}MIIT Key Laboratory of Pattern Analysis and Machine Intelligence, \\
    College of Computer Science and Technology, \\
    Nanjing University of Aeronautics and Astronautics\\
    
    \{haodong\_chen,xiang\_chen\}@nuaa.edu.cn
}

\usepackage{newfloat}
\usepackage{listings}
\DeclareCaptionStyle{ruled}{labelfont=normalfont,labelsep=colon,strut=off} 
\floatstyle{ruled}
\newfloat{listing}{tb}{lst}{}
\floatname{listing}{Listing}

\usepackage{booktabs}

\usepackage{amsmath}
\usepackage{amssymb}

\title{Knowing but Not Saying: Preventing Factual Access \\ Failures in LLM SFT via Recall-Anchored Distillation
}

\begin{document}

\maketitle

\begin{abstract}
Supervised fine-tuning (SFT) can degrade factual behavior outside the target domain. This degradation is often described as catastrophic forgetting, yet open-ended factual failures do not necessarily imply that the underlying facts have been erased. In this work, we identify a more specific phenomenon, factual access failure: after domain SFT, models can still recognize or rank the correct answer under constrained evaluation, while failing to produce it in closed-book generation. Through benchmark-level comparisons, same-fact multiple-choice and generation probes, and failure-mode analysis, we show that SFT-induced factual degradation reflects both genuine wrong-answer generations and expression-level failures such as verbosity, formatting mismatch, and exact-match artifacts. To address this problem, we introduce Recall-Anchored Distillation (RAD), a base-anchored self-distillation objective that preserves out-of-distribution generation behavior by aligning the adapted model with the original base model's soft continuation distribution on unlabeled OOD text. RAD requires no gold OOD answers, external judges, or labeled factual data. Across three backbones fine-tuned on MedMCQA, RAD recovers a consistent portion of the lost OOD recall while preserving target-domain adaptation. Compared with replay on the same OOD text, RAD shows that the key preservation signal is the base model's soft distribution rather than additional text exposure alone.
\end{abstract}


\section{Introduction}

Supervised fine-tuning (SFT) is a standard approach for adapting large language models (LLMs) to specific tasks, domains, and output formats, including instruction following, domain-specific question answering, and structured response generation~\cite{ouyang2022traininglanguagemodelsfollow,
chung2022scalinginstructionfinetunedlanguagemodels,
hu2021loralowrankadaptationlarge,
zhang2025instructiontuninglargelanguage,
yang2024unveilinggeneralizationpowerfinetuned}. In high-stakes domains such as medicine, this adaptation is often necessary because a base model must learn the format, terminology, and decision boundaries of the target task~\cite{pal2022medmcqalargescalemultisubject,
singhal2023expertlevelmedicalquestionanswering,
savage2024finetuninglargelanguage,
li-etal-2024-llamacare}. Yet SFT is not neutral: it can alter the topic preferences, stylistic behavior, factual behavior, and use of pretraining knowledge of a model~\cite{zhang2024iclr-dissecting,gekhman2024doesfinetuningllmsnew,kotha2024understandingcatastrophicforgettinglanguage,
Wang2024RealTime,
yang2024unveilinggeneralizationpowerfinetuned,
li2025preservingdiversitysupervisedfinetuning,
Ye_2025}. Thus, a fine-tuned model may improve on the training task while losing out-of-domain factual reliability.

A decline in out-of-distribution (OOD) factual performance following SFT is frequently attributed to catastrophic forgetting or factual degradation~\cite{luo2025empiricalstudycatastrophicforgetting,
li2024revisitingcatastrophicforgettinglarge,
zhang2024iclr-dissecting,
wu2025mitigatingforgettingllmfinetuning}. However, reduced exact-match accuracy during open-ended generation does not conclusively demonstrate fact erasure. The model may merely lose access to the information, recognize correct answers only from candidates, or produce semantically correct responses that fail to satisfy strict evaluation criteria. Recent literature on spurious forgetting indicates that such performance decreases often reflect shifts in alignment or elicitation strategies rather than a genuine loss of stored knowledge~\cite{zheng2025spuriousforgettingcontinuallearning}. Genuinely erased facts require restoration, whereas recognizable yet unreliable facts indicate failures in knowledge access or output expression.

To distinguish these possibilities, we compare classification-style evaluation
with open-ended generation, following recent concerns that different answer
formats probe different aspects of LLM knowledge and behavior
~\cite{tan-etal-2025-uaqfact,
chen2023factualitycomprehensiveevaluationlarge,
elhady2025wickedsimplemethodmake,
rahmani2025selfcorrectinglargelanguagemodels}. Classification-style probes test whether the model can select or rank the correct answer from candidates, whereas open-ended generation requires producing the answer directly. We find a clear dissociation after domain SFT: recognition-style performance remains comparatively stable on several OOD benchmarks, while open-ended factual generation declines sharply. A paired evaluation of the same facts under multiple-choice and open-ended formats further shows that many facts remain selectable but are no longer generated correctly. Failure-mode analysis reveals both genuinely incorrect answers and expression-level failures, including excessive verbosity, formatting mismatch, and exact-match artifacts~\cite{NEURIPS2023_f323d594,
wang2024factualitylargelanguagemodels,
li2025preservingdiversitysupervisedfinetuning}. We call this gap between retained factual capability and failed open-ended generation factual access failure, with expression failures as a major subclass.

\begin{figure*}[t] 
    \centering 
    \includegraphics[width=0.95\linewidth]{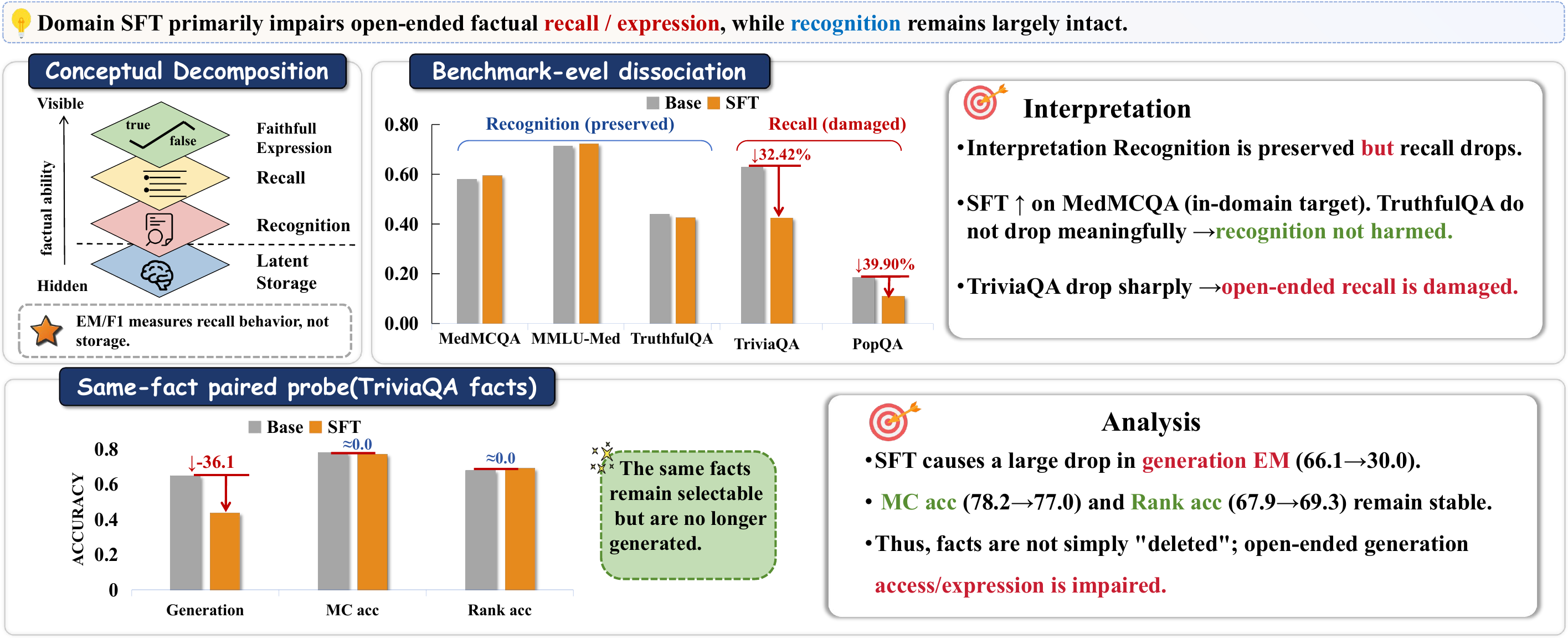} 
    \caption{ Diagnostic overview of factual access failures. Panel (a) separates latent factual storage from observable behaviors: recognition, recall, and expression. Panel (b) shows the benchmark-level comparison in Finding~1. Panel (c) shows the same-fact diagnosis in Finding~2. } 
    \label{fig:diagnosis} 
\end{figure*}

Motivated by this diagnosis and by prior work on mitigating forgetting through
pretraining-simulation, replay, or reference-model regularization
~\cite{chen2020recalllearnfinetuningdeep,
https://doi.org/10.1002/adma.202313608,
chen2026onpolicyreplaycontinualsupervised},
we introduce Recall-Anchored Distillation (RAD), a base-anchored
self-distillation objective for domain adaptation. RAD combines the standard supervised target-domain loss with an unlabeled OOD anchoring stream. For each OOD prefix--continuation example, the base model provides a soft next-token distribution, and the adapted model is trained to match it on continuation tokens. This distributional, rather than answer-based, anchor requires no gold OOD answers, rationales, external judges, or labeled factual data. Using the base model's continuation distribution as a reference~\cite{hinton2015distillingknowledgeneuralnetwork}, RAD limits drift from pre-adaptation OOD generation behavior while allowing target-domain learning. Our contributions are as follows:
\begin{itemize}
    \item We identify factual access failure as a diagnostic framework for SFT-induced degradation in open-ended OOD factual generation, supported by three complementary analyses.

    \item We propose Recall-Anchored Distillation (RAD), a base-anchored self-distillation method that preserves factual access by aligning to the base model's soft OOD continuation distribution.

    \item We evaluate RAD on three backbones fine-tuned on MedMCQA, showing that it recovers OOD factual generation while preserving in-domain gains and outperforming baselines.
\end{itemize}

\section{Related Work}

\subsection{SFT-Induced Factual Degradation in LLMs}

Recent work shows that such adaptation can unintentionally degrade factual reliability \citep{gekhman2024doesfinetuningllmsnew,kaplan2026finetuningencourageshallucinationsfix,gong2025parameterspromptsunderstandingmitigating}. In factual knowledge injection, Gekhman et al. show that knowledge unsupported by the base model is learned more slowly and can increase hallucinations once learned \citep{gekhman2024doesfinetuningllmsnew}; Kang et al. show that unfamiliar fine-tuning examples can shape hallucinated prediction forms \citep{kang2024unfamiliarfinetuningexamplescontrol}; and Zucchet et al. study factual-recall learning dynamics during fine-tuning \citep{zucchet2025languagemodelslearnfacts}. Complementary work on factual QA fine-tuning studies factually correct supervision, showing that fine-tuning can still alter the extraction of pretrained factual associations: models may over-amplify task-specific shortcuts, exhibit frequency shocks, or rely on relation-level cues while ignoring subject-specific information \citep{kazemi2023understandingfinetuningfactualknowledge,ghosal2024understandingfinetuningfactualknowledge,gong2025parameterspromptsunderstandingmitigating}. Together, these studies show that SFT can degrade factual behavior even without noisy supervision. However, they leave open a finer-grained question: when closed-book factual accuracy drops after SFT, has the model lost the underlying fact, failed to access a still-recognizable fact in open-ended generation, or expressed the answer in a surface form that closed-book metrics cannot reliably match?

\subsection{Hidden Knowledge and Recall Failures}

Representation engineering studies how model behaviors and concepts are encoded in internal representations and how these representations can be analyzed or manipulated to control model outputs. Studies of implicit inference argue that fine-tuning can shift the model's inferred task distribution toward fine-tuning data, suppressing pretrained capabilities recoverable under alternative elicitation conditions \citep{kotha2024understandingcatastrophicforgettinglanguage}. Work on spurious forgetting similarly shows that apparent forgetting can result from disrupted task alignment rather than true knowledge loss \citep{zheng2025spuriousforgettingcontinuallearning}. Complementary factual probing studies further show that models may encode more information than direct generation reveals, and that ranking-based, internal, or alternative probes can recover factual knowledge missed by standard closed-book generation \citep{gekhman2025insideouthiddenfactualknowledge, orgad2025llmsknowshowintrinsic}. These findings motivate our focus on factual access. Rather than assuming that SFT-induced generation failures reflect storage loss, we ask whether the same facts remain recognizable under constrained probes and where extraction breaks down. Our repair strategy relates closely to knowledge distillation and KL-constrained training, which stabilize learning by matching a teacher or reference distribution \citep{hinton2015distillingknowledgeneuralnetwork, ziegler2020finetuninglanguagemodelshuman, stiennon2022learningsummarizehumanfeedback, ouyang2022traininglanguagemodelsfollow}. Unlike generic KL constraints that keep an adapted policy close to a reference model, our method uses the base model as a teacher only on OOD factual continuations, namely general-domain factual continuations outside the SFT target domain, and regularizes the adapted model toward the base model's soft token distribution while optimizing target-domain SFT behavior.

\section{Preliminary Studies}

\label{sec:prelim_diagnosis}

\begin{figure}[t]
    \centering
    \includegraphics[width=\linewidth]{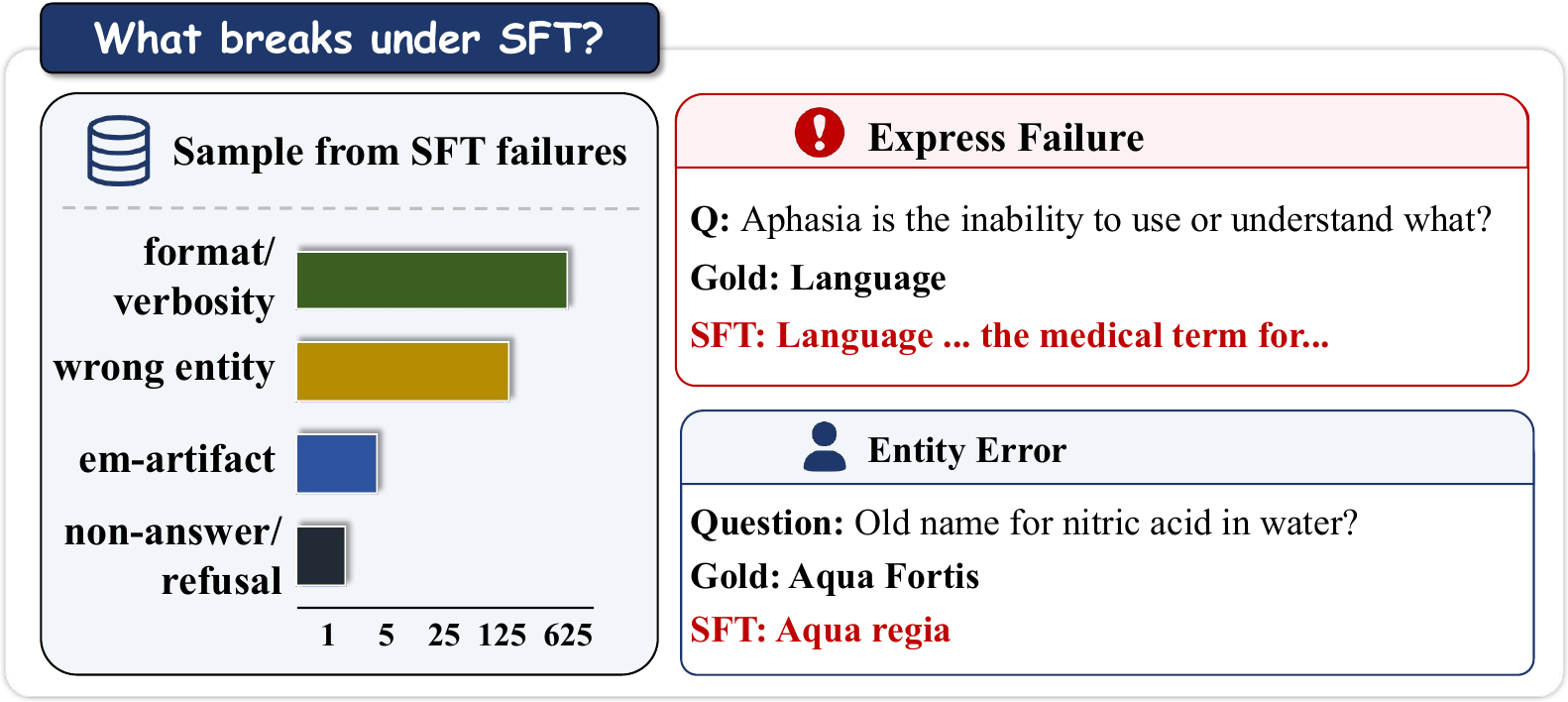}
    \caption{
    Failure modes among Base-correct/SFT-wrong examples. Most recall failures
    are format or verbosity mismatches rather than clean wrong-entity errors.
   }
    \label{fig:failure_modes}
\end{figure}
This section diagnoses factual degradation after domain SFT. A drop in closed-book generation accuracy shows failure to produce the correct answer, but not erasure of the corresponding fact. Consistent with recent studies on spurious forgetting and the distinction between knowledge encoding and recall \citep{calderon2026shelveslostkeysrecall,gekhman2025insideouthiddenfactualknowledge,kotha2024understandingcatastrophicforgettinglanguage,zheng2025spuriousforgettingcontinuallearning}, apparent degradation can reflect task-alignment shifts rather than loss of underlying capabilities. We distinguish memory erasure from expression failure through \textbf{three findings}.

\subsection{F1: Models Recognize What They Cannot Recall}
We compare recognition and recall at the benchmark level because domain SFT may broadly degrade factual behavior or selectively impair fact elicitation. Recognition tests whether the model selects correct answers under constrained evaluation, whereas recall tests whether it generates them without candidate support. We measure recognition with multiple-choice evaluation and recall with closed-book QA metrics, EM/F1~\citep{roberts-etal-2020-much}. Domain SFT improves the target-domain QA task, but this gain coincides with substantial degradation in open-ended factual recall on held-out QA benchmarks (see Figure~\ref{fig:diagnosis}b): for example, TriviaQA EM drops from 65.03 to 43.95 on the full benchmark, while MedMCQA and MMLU remain stable or slightly improve. This pattern is not yet a same-fact dissociation because datasets differ in content and format; instead, it motivates a controlled diagnosis of whether the same facts remain accessible under constrained evaluation.

\subsection{F2: Facts Are Stored, but Inaccessible}
\label{sec:paired}

To separate elicitation format from factual content, we evaluate recognition and recall on matched facts. We denote the factual association targeted by a QA item as $f=(s,r,o)$, where $s$, $r$, and $o$ represent subject, relation, and answer object \citep{petroni-etal-2019-language,ghosal2024understandingfinetuningfactualknowledge}; this is an analytic notation rather than a claim that TriviaQA items form explicit knowledge graphs. For each item, we pair the original open-ended question with recognition-style probes for the same answer object. On this subset, SFT substantially reduces open-ended generation (see Figure~\ref{fig:diagnosis}c): TriviaQA EM drops from 63.1 to 43.8, with marginal changes in multiple-choice accuracy and no ranking degradation. Under teacher forcing, the gold answer remains highly ranked: its first token stays near the top, with the top-1 rate nearly unchanged from Base to SFT. Thus, unreliably generated answers remain selectable or highly preferred under constrained or teacher-forced evaluation. We therefore use factual access failure as a behavioral diagnosis: the model selectively fails to produce certain facts in open-ended generation, rather than uniformly losing factual behavior.

\subsection{F3: Recall Failures Are Largely Expression Failures}
\label{sec:whatbreaks}

The paired probe shows that recognition and recall can diverge, motivating our analysis of these recall failures in generated text. We analyze 300 TriviaQA examples where the base model is EM-correct but the SFT model is EM-wrong. To distinguish factual errors from surface-form expression failures, we label each SFT output as a format/verbosity mismatch, wrong entity, EM artifact (semantically valid but rejected by exact match), or non-answer/refusal. The labels agree with an independent LLM judge (DeepSeek; In Appendix) on 97.3\% of cases. As Figure~\ref{fig:failure_modes} shows, format/verbosity mismatch dominates, accounting for 77.3\% of the analyzed cases. This suggests that expression failures explain most Base-correct/SFT-wrong errors, although a non-negligible subset still corresponds to genuine wrong-answer generations. As a classifier-independent elicitation check, a simple 4-shot prompt recovers 90.0\% of the same cases, indicating that many failures in this sample remain recoverable under different prompting rather than reflecting irreversible knowledge erasure. Together, these findings motivate a method that preserves open-ended factual expression during domain SFT.

\begin{figure*}[t]
\centering
\includegraphics[width=0.9\linewidth]{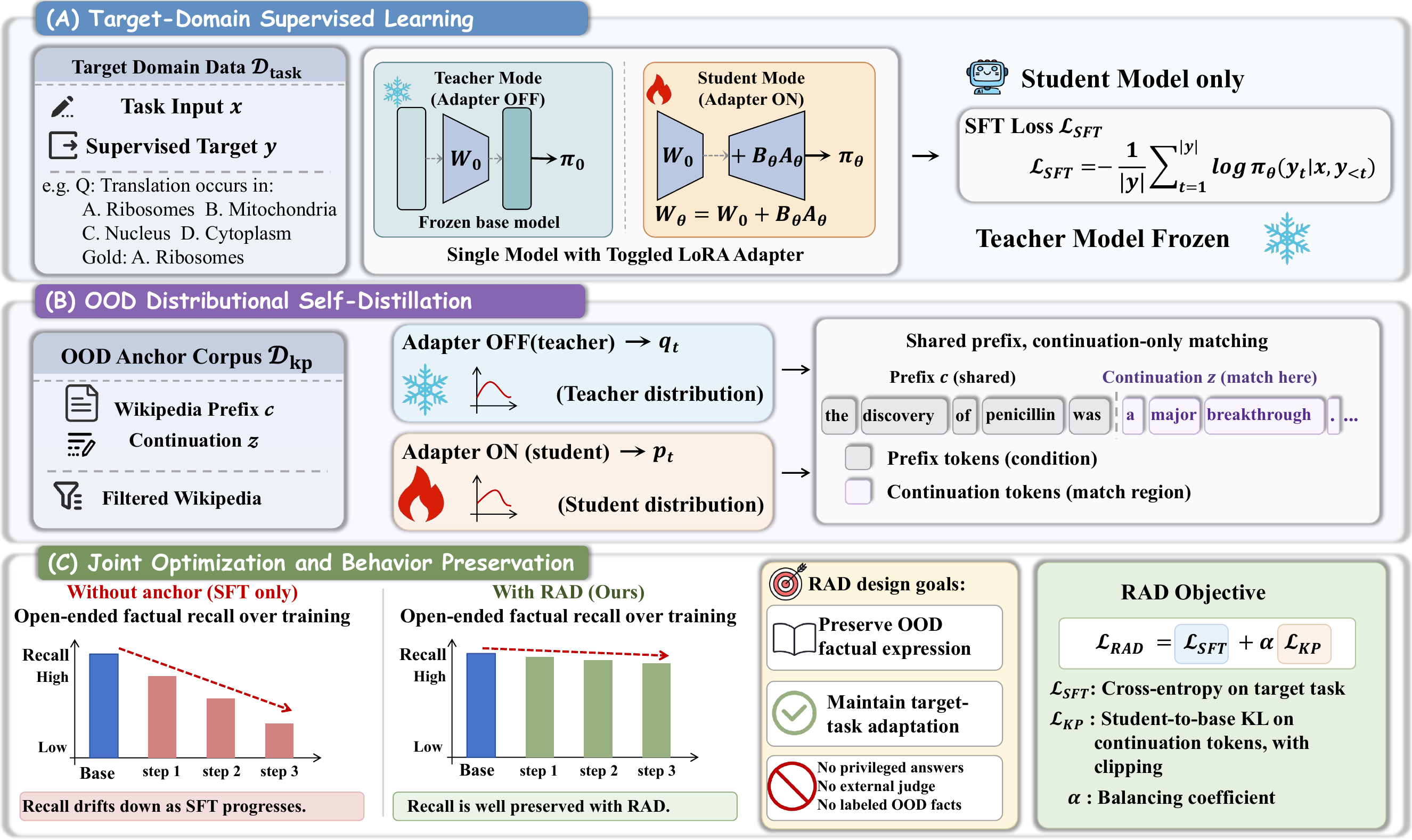}
\caption{
Overview of Recall-Anchored Distillation (RAD).
}
\label{fig:Method}
\end{figure*}

\section{Methodology}
\label{sec:method}

Motivated by the diagnosis in Section~\ref{sec:prelim_diagnosis}Preliminary, we introduce Recall-Anchored Distillation (RAD), a base-anchored self-distillation objective designed to preserve out-of-distribution (OOD) generation behavior during domain adaptation. Here, OOD is defined relative to the target domain. Figure~\ref{fig:Method} illustrates the overall RAD design.

\subsection{Teacher and Student from One Model}

Let \(f_{\Theta_0}\) denote the original base model, which induces the next-token distribution \(\pi_0(\cdot \mid u)=\mathrm{softmax}(f_{\Theta_0}(u))\) for input prefix \(u\). We adapt it with Low-Rank Adaptation (LoRA)~\citep{hu2021loralowrankadaptationlarge}: each frozen weight matrix \(W_0\in\Theta_0\) receives a trainable low-rank update \(W_\theta = W_0 + B_\theta A_\theta\), yielding the adapted student distribution \(\pi_\theta(\cdot \mid u)\). RAD implements teacher and student in one model through the LoRA switch: adapter-off mode outputs the base distribution \(\pi_0\) as the teacher, and adapter-on mode outputs \(\pi_\theta\) as the student. Thus, the teacher is not an external oracle but the original base model itself, anchoring adaptation to the pre-adaptation generation behavior of the base model and following the broader use of teacher distributions to preserve model behavior during adaptation~\citep{hinton2015distillingknowledgeneuralnetwork,li2017learningforgetting}.

\begin{table*}[t]
\centering
\small
\setlength{\tabcolsep}{4.5pt}
\begin{tabular*}{\textwidth}{@{\extracolsep{\fill}}l c ccc cc cc}
\toprule
 & \textbf{In-Domain} $\uparrow$ &
 \multicolumn{3}{c}{\textbf{OOD Recognition} (multiple choice) $\uparrow$} &
 \multicolumn{4}{c}{\textbf{OOD Recall} (closed-book) $\uparrow$} \\
\cmidrule(lr){2-2}\cmidrule(lr){3-5}\cmidrule(lr){6-9}
 & MedMCQA & MMLU-Med & MMLU-Other & TruthfulQA &
 \multicolumn{2}{c}{TriviaQA} & \multicolumn{2}{c}{PopQA} \\
\cmidrule(lr){6-7}\cmidrule(lr){8-9}
\textbf{Method} & Acc & Acc & Acc & MC2 & EM & F1 & EM & F1 \\
\midrule
\multicolumn{9}{l}{\emph{Llama-3.1-8B}} \\
\quad Base & 58.14 & 71.54 & 64.82 & 44.17 & 65.03 & 71.04 & 19.30 & 23.67 \\
\quad + Standard SFT & 59.43 & \underline{72.20} & \textbf{65.51} & \textbf{42.65} & \underline{43.95} & \underline{56.59} & \underline{11.60} & \underline{16.03} \\
\quad + Replay & \underline{59.53} & 71.90 & 64.88 & 41.48 & 32.87 & 47.80 & 3.40 & 8.06 \\
\quad + RAD (Ours) & \textbf{60.05} & \textbf{72.57} & \underline{65.24} & \underline{42.07} & \textbf{51.62} & \textbf{62.16} & \textbf{15.20} & \textbf{19.63} \\
\midrule
\multicolumn{9}{l}{\emph{Qwen2.5-7B-Instruct}} \\
\quad Base & 56.18 & 76.45 & 73.26 & 64.75 & 32.96 & 43.06 & 5.51 & 11.18 \\
\quad + Standard SFT & 61.27 & 76.04 & \textbf{73.75} & 49.03 & \underline{23.27} & \textbf{31.41} & \underline{4.95} & \underline{8.77} \\
\quad + Replay & \underline{61.68} & \underline{77.06} & 73.47 & \underline{50.21} & 16.09 & 25.31 & 1.77 & 6.11 \\
\quad + RAD (Ours) & \textbf{61.99} & \textbf{77.18} & \underline{73.68} & \textbf{58.14} & \textbf{23.68} & \underline{30.60} & \textbf{6.81} & \textbf{9.99} \\
\midrule
\multicolumn{9}{l}{\emph{Qwen2.5-3B-Instruct}} \\
\quad Base & 51.54 & 68.87 & 66.74 & 58.74 & 30.57 & 39.10 & 2.97 & 8.01 \\
\quad + Standard SFT & \textbf{55.75} & 67.97 & \underline{67.45} & \underline{45.89} & 1.04 & \underline{13.58} & \underline{1.45} & 5.17 \\
\quad + Replay & 55.22 & \textbf{68.99} & 67.15 & 44.99 & \underline{3.39} & 13.24 & 0.98 & \underline{5.38} \\
\quad + RAD (Ours) & \underline{55.41} & \underline{68.83} & \textbf{68.12} & \textbf{54.88} & \textbf{9.04} & \textbf{22.00} & \textbf{2.66} & \textbf{6.77} \\
\bottomrule
\end{tabular*}
\caption{
Main results across three backbones on in-domain MedMCQA, OOD recognition, and OOD closed-book recall.
}
\label{tab:main}
\end{table*}

\subsection{Training Streams}

\paragraph{Target-domain Supervision.}
The primary training signal comes from the supervised target-domain dataset \(\mathcal{D}_{\mathrm{task}}=\{(x,y)\}\), where \(x\) denotes the task input and \(y\) denotes the supervised target. This term is identical to standard domain SFT: it teaches the adapted model to solve the target-domain task and is the only component of RAD that uses task labels. For a mini-batch \(\mathcal{B}_{\mathrm{task}}\subset\mathcal{D}_{\mathrm{task}}\), the adapter-on student is optimized with the teacher-forced cross-entropy objective:
\begin{equation}
\label{eq:sft_loss}
\mathcal{L}_{\mathrm{SFT}}
=
-\frac{1}{|\mathcal{B}_{\mathrm{task}}|}
\sum_{(x,y)\in\mathcal{B}_{\mathrm{task}}}
\frac{1}{|y|}
\sum_{t=1}^{|y|}
\log \pi_\theta(y_t \mid x,y_{<t}).
\end{equation}
This loss drives domain adaptation by updating only LoRA parameters while keeping the base model frozen.

\paragraph{OOD Anchor Construction.}
The secondary training stream is an unlabeled OOD anchor corpus \(\mathcal{D}_{\mathrm{kp}}=\{(c,z)\}\), where \(c\) denotes a Wikipedia prefix and \(z\) its natural continuation. The continuation \(z\) is not treated as a hard label. Instead, it specifies the token positions over which the adapted model aligns with the base model next-token distribution, consistent with soft teacher distributions in knowledge distillation~\citep{hinton2015distillingknowledgeneuralnetwork}. Since the diagnosis attributes the degradation primarily to OOD open-ended generation rather than target-domain learning itself, RAD sources this anchor stream from generic OOD text instead of target-domain training data.

To mitigate potential benchmark contamination, we filter the anchor corpus via case-insensitive substring matching, discarding any prefix--continuation pair that contains evaluation questions, answer strings, or normalized aliases from benchmarks including PopQA and TriviaQA. For each retained anchor example, we concatenate the prefix and continuation into a single sequence \(s=[c;z]\). The prefix provides contextual grounding, while the continuation designates the matching region. We define a binary continuation mask \(m_t\in\{0,1\}\), where \(m_t=1\) indicates that the prediction target \(s_t\) falls within \(z\), and \(m_t=0\) for prefix positions.

\subsection{Distributional Anchoring Objective}

RAD processes each OOD sequence in two model modes. With the adapter disabled, the adapter-off teacher provides the base next-token distribution:
\begin{equation}
\label{eq:teacher_dist}
q_t
=
\pi_0(\cdot \mid s_{<t}).
\end{equation}
With the adapter enabled, the adapter-on student provides the adapted next-token distribution:
\begin{equation}
\label{eq:student_dist}
p_t
=
\pi_\theta(\cdot \mid s_{<t}).
\end{equation}
RAD aligns the student to the frozen base distribution using reverse KL, with the optimized student distribution as the first argument and the base distribution as the reference:
\begin{equation}
\label{eq:token_kl}
d_t
=
D_{\mathrm{KL}}
\left(
p_t
\parallel
q_t
\right).
\end{equation}
This direction discourages the adapted model from assigning probability mass to OOD continuations that the base model deems improbable. To prevent a few highly divergent tokens from dominating the objective, we clip the per-token penalty at threshold \(\tau\). The OOD distillation loss is:
\begin{equation}
\label{eq:kp_loss}
\mathcal{L}_{\mathrm{KP}}
=
\frac{1}{\sum_t m_t}
\sum_t
m_t
\min\!\left(d_t,\tau\right).
\end{equation}
The mask ensures that prefix tokens condition both model modes but contribute no gradient through the distillation loss.

\subsection{Joint Optimization and Implementation}

\paragraph{Joint Objective.}
RAD jointly optimizes target-domain adaptation and OOD distributional self-distillation via:
\begin{equation}
\label{eq:rad_loss}
\mathcal{L}_{\mathrm{RAD}}
=
\mathcal{L}_{\mathrm{SFT}}
+
\alpha\mathcal{L}_{\mathrm{KP}},
\end{equation}
where \(\alpha\) controls the strength of the OOD anchor. We set \(\alpha=1.5\) and \(\tau=5.0\) as default values. The supervised loss drives target-domain learning, while the OOD distillation loss discourages the adapted model from drifting away from the base model on generic continuations. RAD therefore does not freeze or constrain the adapter globally; it allows target-task adaptation while selectively anchoring the continuation distribution on OOD text.

\paragraph{Single-model Implementation.}
RAD is memory-efficient because it requires only one model instance in memory. For target-domain batches, we enable the LoRA adapter and compute \(\mathcal{L}_{\mathrm{SFT}}\). For OOD anchor batches, we run the same sequence twice: first with the adapter disabled under \texttt{no\_grad} to obtain teacher logits, and then with the adapter enabled to obtain student logits for computing \(\mathcal{L}_{\mathrm{KP}}\). Only the adapter-on student pass contributes gradients. This implementation avoids loading a separate teacher model and makes the teacher-student distinction a mode switch of the same backbone. The additional cost is one extra forward pass per OOD anchor batch for teacher-logit computation. Since RAD pairs each target-domain optimizer step with one OOD anchor batch, the method adds modest training-time overhead while preserving the same inference-time architecture as standard LoRA SFT.

\section{Experiments}
\label{sec:results}

\begin{figure*}[t]
    \centering
    \includegraphics[width=0.9\linewidth]{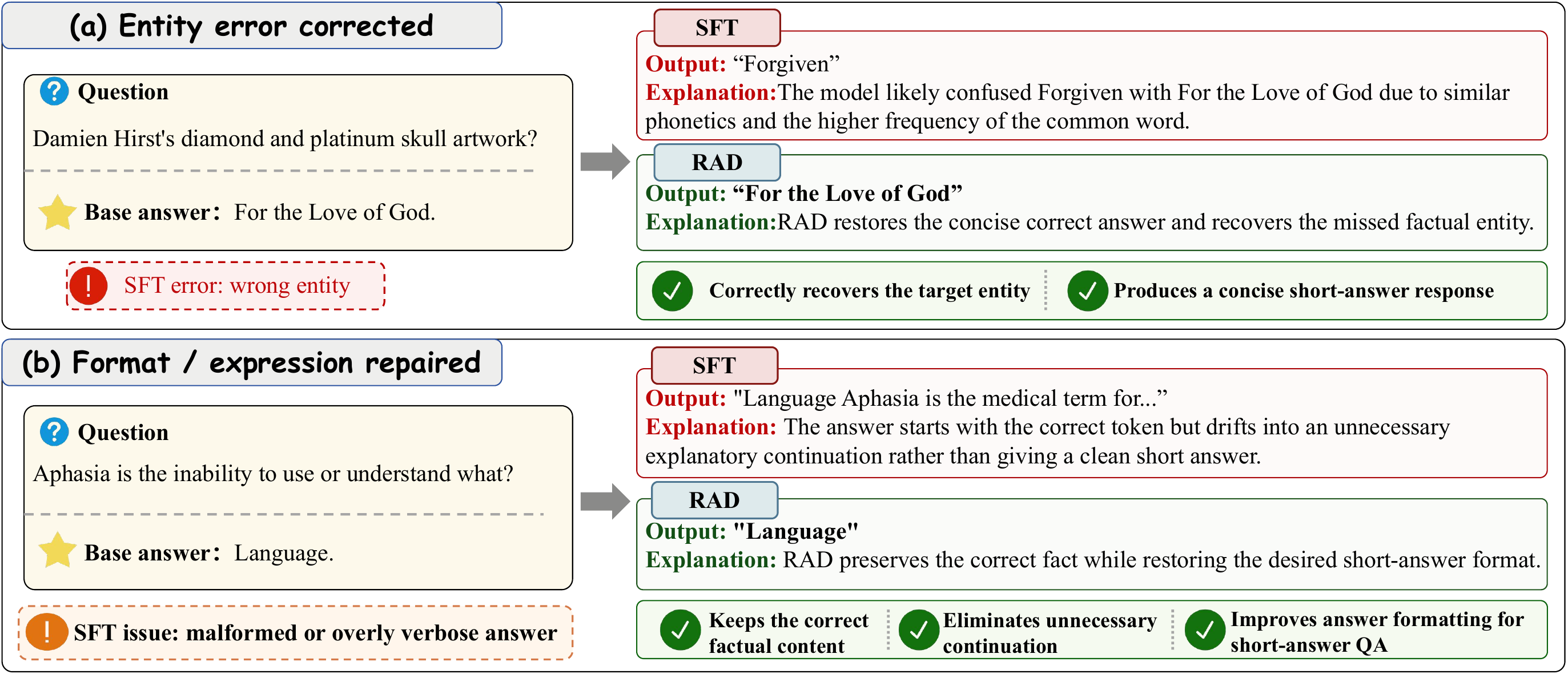}
    \caption{
   Qualitative examples showing how RAD repairs open-ended factual access failures on TriviaQA.
    }
    \label{fig:case_study}
\end{figure*}

\subsection{Experimental Setup}


\paragraph{Training Setup.}
We evaluate three backbones with different sizes and instruction-tuning properties: Llama-3.1-8B, Qwen2.5-7B-Instruct, and Qwen2.5-3B-Instruct~\citep{qwen2025qwen25technicalreport}. For Llama-3.1-8B, we use the base checkpoint rather than the instruction-tuned variant. All adapted systems use LoRA adapters~\citep{hu2021loralowrankadaptationlarge} with the pretrained backbone frozen, and all methods share the same target-domain data and evaluation protocol. RAD defaults to reverse KL ($\alpha=1.5$, $\tau=5.0$), with full details in Appendix.

\paragraph{Training Data.}
We use MedMCQA~\citep{pal2022medmcqalargescalemultisubject} as the target-domain fine-tuning task, training adapted models on the training split and evaluating them on the validation split. Each example is a four-choice QA instance whose target response contains the correct option and its explanation. For the OOD anchor stream, RAD uses an unlabeled Wikipedia-style corpus, following the use of Wikipedia-derived corpora in language modeling benchmarks~\citep{merity2016pointersentinelmixturemodels}. Here OOD is defined relative to the target-domain task. The anchor corpus contains 10,000 prefix--continuation examples, where the prefix provides context and the continuation marks the tokens for distributional anchoring. We filter this corpus against PopQA and TriviaQA entities and aliases to reduce benchmark contamination.

\paragraph{Compared Baselines.}
We compare RAD against three systems. \textbf{Base} is the original backbone
without task adaptation and serves as a no-adaptation reference. \textbf{Standard
SFT} fine-tunes the model only on MedMCQA, following the standard supervised
fine-tuning paradigm used for adapting language models to target behaviors or
tasks~\citep{ouyang2022traininglanguagemodelsfollow}. \textbf{Replay} trains on the same amount of
OOD anchor text as RAD, but uses a hard-label language-modeling loss rather than
matching the base model's next-token distribution. This baseline controls for
whether the benefit comes simply from adding general-domain text.


\subsection{Benchmarks and Metrics}

We evaluate target-domain adaptation on MedMCQA~\citep{pal2022medmcqalargescalemultisubject} and report validation accuracy. To assess recognition-style OOD behavior, we use MMLU~\citep{hendrycks2021measuringmassivemultitasklanguage}, separating medical subjects as MMLU-Med and non-medical subjects as MMLU-Other, together with TruthfulQA-MC2~\citep{lin2022truthfulqameasuringmodelsmimic}. For open-ended OOD factual generation, we evaluate TriviaQA~\citep{joshi2017triviaqalargescaledistantly} and PopQA~\citep{mallen2023trustlanguagemodelsinvestigating}, reporting exact match (EM) and token-level F1 over acceptable aliases. All multiple-choice scoring, generation settings, normalization rules, and additional generation-based multiple-choice checks are detailed in Appendix.

\subsection{Main Results}
\subsubsection{For Q1: Does Domain SFT Cause Capability Collapse or Selective OOD Shift?}

Table~\ref{tab:main} shows domain SFT induces a selective OOD shift, not uniform capability degradation. Across three backbones, standard SFT improves in-domain MedMCQA performance, confirming effective target-domain adaptation. Recognition-style OOD performance remains comparable on MMLU-Med and MMLU-Other, suggesting domain tuning does not simply reduce general multiple-choice competence. Degradation is largest in open-ended or truthfulness-oriented OOD behavior. For \texttt{Llama-3.1-8B}, TriviaQA EM drops from 65.03 to 43.95, and for \texttt{Qwen2.5-3B-Instruct} from 30.57 to 1.04. \texttt{Qwen2.5-7B-Instruct} shows a related pattern, with TruthfulQA-MC2 dropping from 64.75 to 49.03. These results suggest that domain SFT primarily perturbs pre-adaptation knowledge expression under OOD conditions rather than uniformly erasing recognition-style ability.

\subsubsection{For Q2: Can RAD Recover OOD Factual Behavior While Preserving Target-Domain Adaptation?}

RAD mitigates the SFT-induced OOD shift while retaining competitive in-domain performance. On \texttt{Llama-3.1-8B}, RAD improves both TriviaQA and PopQA EM, while achieving the best MedMCQA accuracy among fine-tuned systems. The same trend extends to the Qwen backbones, though the affected OOD axis differs by model: RAD substantially recovers TruthfulQA-MC2 on \texttt{Qwen2.5-7B} and TriviaQA EM on \texttt{Qwen2.5-3B}, with only a small MedMCQA drop for the latter. These results suggest that anchoring the adapted model to the base model on unlabeled OOD continuations can partly recover damaged generation behavior without materially undermining target-domain learning. Since RAD introduces no annotated OOD facts, the improvement is better viewed as preservation of generation behavior rather than re-learning the evaluation answers.

\subsubsection{For Q3: Is the Soft Base Distribution Necessary Beyond Same-Text Replay?}

Replay isolates the role of the soft base distribution: it uses the same OOD anchor corpus as RAD but replaces distributional anchoring with hard-label language modeling. If RAD gained only from extra general-domain text, Replay would provide comparable preservation. Instead, Replay lags RAD on OOD recall and often worsens SFT degradation. This is clearest for \texttt{Llama-3.1-8B}, where Replay reduces PopQA EM to 3.40, versus 11.60 under standard SFT and 15.20 under RAD. Thus, the key signal is not anchor text alone but the token-level base-model distribution over that text. Matching this soft distribution preserves relative preferences over OOD continuations, whereas hard-label replay adds a competing objective that does not directly protect generation behavior disrupted by domain SFT.

\begin{table}[t]
\centering
\small
\setlength{\tabcolsep}{4.5pt}
\begin{tabular}{lccc}
\toprule
Variant & MedMCQA & TriviaQA EM & PopQA EM \\
\midrule
Standard SFT & 59.43 & 43.95 & 11.60 \\
RAD default & 60.05 & 51.62 & 15.20 \\
\midrule
Forward KL & 59.89 & 52.88 & 16.74 \\
Symmetric KL & 59.93 & 52.57 & 16.58 \\
\midrule
No clipping & 59.98 & 51.59 & 16.58 \\
\(\tau=2.5\) & 59.98 & 52.17 & 16.77 \\
\(\tau=10.0\) & 60.12 & 51.84 & 16.60 \\
\midrule
On-policy anchor & 60.24 & 51.44 & 15.10 \\
\bottomrule
\end{tabular}
\caption{
Ablation results of RAD.
}
\label{tab:rad_ablations}
\end{table}


\begin{figure}[t]
    \centering
    \includegraphics[width=\linewidth]{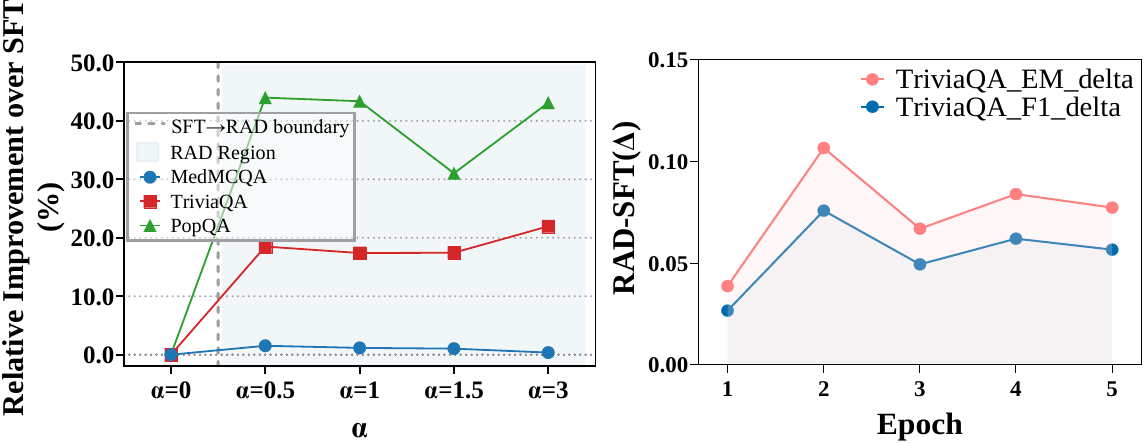}
    \caption{
    Ablation and checkpoint-dynamics results on \texttt{Llama-3.1-8B}. Left: effect of anchoring strength \(\alpha\). Right: TriviaQA accuracy gains of RAD over Standard SFT at each training epoch.
    }
    \label{fig:epoch_alpha}
\end{figure}


\subsubsection{Ablation Analysis.}
We conduct an ablation study of RAD on Llama-3.1-8B to evaluate the impact of anchoring strength, checkpoint dynamics, KL divergence direction, clipping thresholds, and sampling strategies. Figure~\ref{fig:epoch_alpha} demonstrates that the performance improvements are driven by positive OOD anchoring. Compared to the standard SFT baseline ($\alpha=0$), setting $\alpha>0$ enhances performance on TriviaQA and PopQA while maintaining accuracy on MedMCQA near the baseline level. The observed non-monotonic gains indicate that the parameter $\alpha$ regulates the degree of OOD preservation rather than converging to a universal optimum; consequently, we adopt $\alpha=1.5$ as a stable default configuration. Across all five training epochs, RAD consistently yields higher accuracy on TriviaQA compared to standard SFT, indicating that the benefits are not specific to the final checkpoint. Furthermore, Table~\ref{tab:rad_ablations} illustrates the robustness of the method across variations in forward KL, symmetric KL, alternative clipping thresholds, and on-policy anchoring, with EM scores exceeding 51\% on TriviaQA and remaining near 60\% on MedMCQA. These findings suggest that performance preservation stems from aligning the adapted model with the soft OOD continuation distribution of the base model, rather than merely from exposure to the anchor text.

\subsubsection{Qualitative Analysis.}
Figure~\ref{fig:case_study} illustrates qualitative examples of failure modes identified in the preliminary diagnosis. In the entity-error case, SFT replaces target answer with an incorrect but plausible entity, whereas RAD recovers the concise factual answer. In the format-error case, SFT starts with the correct token but continues into an unnecessary explanatory completion, producing a malformed short answer; RAD preserves the factual content and restores the desired answer format. These examples serve only as qualitative illustrations; aggregate trends appear in Table~\ref{tab:main}, Table~\ref{tab:rad_ablations} and Figure~\ref{fig:epoch_alpha}.
\begin{figure}[t]
    \centering
    \includegraphics[width=0.8\linewidth]{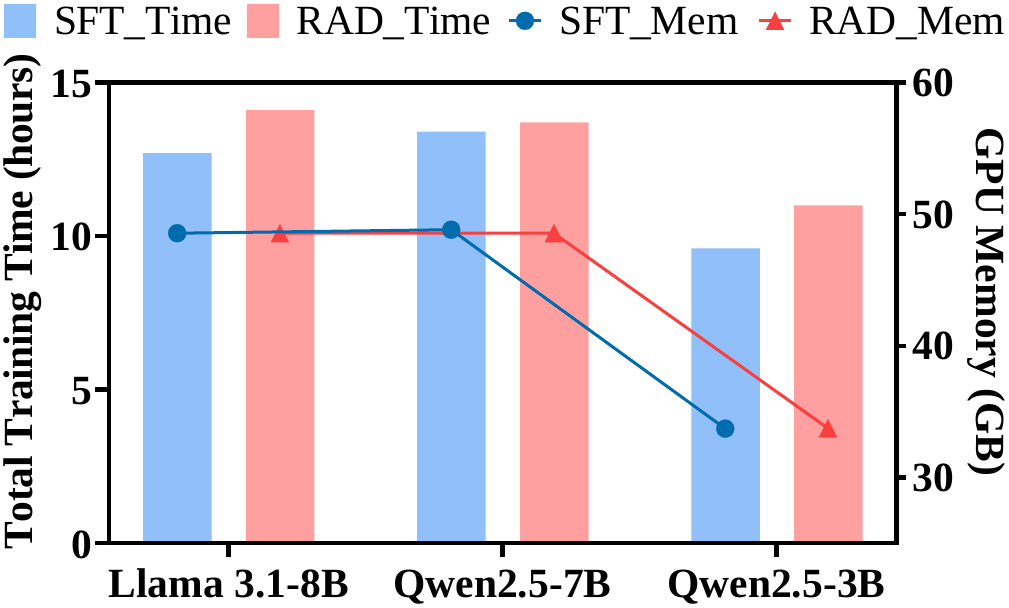}
    \caption{
Training time and GPU memory usage of Standard SFT and RAD across three backbones.
    }
    \label{fig:efficiency}
\end{figure}

\subsubsection{Efficiency Analysis.}
We examine RAD training cost relative to Standard SFT in Figure~\ref{fig:efficiency}. Because RAD uses the same backbone in two LoRA modes instead of a separate teacher model, its memory footprint remains essentially unchanged across all three backbones, with differences within 1 GB. The main overhead is the additional adapter-off forward pass on OOD anchor batches, but the total training-time increase is modest: the largest increase is about 1.4 hours. Overall, RAD adds limited training-time cost while preserving the same inference-time architecture as standard LoRA SFT.

\section{Conclusion}

This work studies factual access failure in domain SFT, where models recognize facts but fail to reliably recall or express them in open-ended OOD generation. Via benchmark comparisons, same-fact probes, and failure-mode analysis, we show SFT-induced degradation is selective: recognition remains stable while open-ended recall and expression degrade. We introduce Recall-Anchored Distillation (RAD), preserving OOD generation by aligning the adapted model with the base model's soft continuation distribution on unlabeled OOD text. Across three MedMCQA-tuned backbones, RAD partially recovers damaged OOD factual behavior while maintaining target-domain adaptation and outperforming replay on identical anchor text.

\bibliography{aaai2027}

\end{document}